\pdfoutput=1

\documentclass[11pt]{article}

\usepackage[]{ACL2023}

\usepackage{times}
\usepackage{array}
\usepackage{longtable}
\usepackage{latexsym}
\usepackage{amsmath}
\usepackage{graphicx}
\usepackage{multirow}
\usepackage{array}
\usepackage{graphicx}
\usepackage{microtype}
\usepackage{amsfonts}
\usepackage{mathrsfs}
\usepackage{multirow}
\usepackage{amssymb}
\usepackage{bbding}
\usepackage{subfigure} 
\usepackage{CJKutf8}
\usepackage{colortbl}
\usepackage{color}
\usepackage{booktabs}
\usepackage{makecell}
\usepackage{calligra}
\usepackage{algorithm} 
\usepackage{subcaption}
\usepackage{subfigure}
\usepackage{algpseudocode}%
\usepackage{algorithmicx}
\usepackage{algpseudocode} 
\usepackage{amsmath}
\usepackage[T1]{fontenc}

\usepackage[utf8]{inputenc}
\usepackage{microtype}
\usepackage{inconsolata}

\title{Large Language Models for Automated Literature Review: An Evaluation of Reference Generation, Abstract Writing, and Review Composition}

\author{
\textbf{Xuemei Tang\textsuperscript{1}}\quad
\textbf{Xufeng Duan\textsuperscript{1}}\quad
\textbf{Zhenguang G. Cai\textsuperscript{1,2}}\\
\textsuperscript{1}Department of Linguistics and Modern Languages, The Chinese University of Hong Kong\\
\textsuperscript{2}Brain and Mind Institute, The Chinese University of Hong Kong\\
}

\begin{document}
\maketitle
\begin{abstract}
Large language models (LLMs) have emerged as a potential solution to automate the complex processes involved in writing literature reviews, such as literature collection, organization, and summarization. However, it is yet unclear how good LLMs are at automating comprehensive and reliable literature reviews. This study introduces a framework to automatically evaluate the performance of LLMs in three key tasks of literature review writing: reference generation, abstract writing, and literature review composition. We introduce multidimensional evaluation metrics that assess the hallucination rates in generated references and measure the semantic coverage and factual consistency of the literature summaries and compositions against human-written counterparts. The experimental results reveal that even the most advanced models still generate hallucinated references, despite recent progress. Moreover, we observe that the performance of different models varies across disciplines when it comes to writing literature reviews. These findings highlight the need for further research and development to improve the reliability of LLMs in automating academic literature reviews. The dataset and code used in this study are publicly available at an anonymous repository~\footnote{\url{https://anonymous.4open.science/r/Eval_LLM_LR-C657}}.
\end{abstract}

\section{Introduction}

The literature review is a critical component of academic writing that aims to synthesize, critique, and assess the current state of knowledge in a particular field. It involves a comprehensive examination of published research articles, theoretical frameworks, and research methodologies related to a specific topic. Conducting a thorough literature review often necessitates extensive reading and summarizing of pertinent literature, which can be a complex and time-consuming process, especially in well-established fields where the number of relevant references can range from dozens to hundreds. To alleviate this burden, researchers have recently turned to advanced deep learning models as a potential tool to aid in the automated generation of literature reviews~\cite{Aliyu_Iqbal_James_2018, Kontonatsios_Spencer_Matthew_Korkontzelos_2020}. 

The emergence of large language models (LLMs) has introduced a promising avenue for automating key aspects of literature review writing, including identifying relevant sources, summarizing findings, and generating coherent syntheses~\cite{Wang_Guo_Yao_Zhang_Zhang_Wu_Zhang_Dai_Zhang_Wen_et_2024, Agarwal_Sahu_Puri_Laradji_Dvijotham_Stanley_Charlin_Pal_2024, hsu-etal-2024-chime}.
While techniques such as Retrieval-Augmented Generation (RAG) can enhance the domain-specific knowledge of LLMs, most researchers rely on vanilla LLMs, such as ChatGPT, for literature review writing without the use of RAG~\cite{Wang_Hu_Wang_Yan_Sheng_He_2024}. Consequently, it is crucial to evaluate the performance of these naive LLMs in the context of literature review writing to determine their effectiveness and limitations.

Therefore, in this paper, we propose a framework for automatically assessing the literature review writing ability of LLMs, using human-written literature reviews as the gold standard and designing metrics for a comprehensive evaluation. We first collect a dataset of human-written literature reviews to serve as a benchmark for evaluating the performance of LLMs. We then ask LLMs to complete three tasks based on the collected dataset: generating references, writing an abstract, and writing a complete literature review based on a given topic. Finally, we evaluate the generated results from several dimensions, including the presence of hallucinations in the references, as well as the semantic coverage and factual consistency of the generated abstract and literature review compared to the human-written context. By assessing the performance of LLMs across these tasks and evaluating their output using our proposed metrics, we aim to provide a comprehensive understanding of their capabilities and limitations in writing literature reviews.

Our contribution can be summarized as follows.

\begin{itemize}
    \item First, we propose a framework for automatically evaluating the literature review writing ability of LLMs, without requiring any human involvement. This framework encompasses multiple stages, including the compilation of a literature review dataset construction, the collection of LLM-generated output, and the evaluation of LLM performance.
    
    \item Second, we collect 1,105 literature reviews from 51 journals across 5 disciplines as the ground truth. We then design three tasks for accessing LLMs in literature writing: reference generation, abstract writing, and literature composition on a given topic.
    
    \item Then, we evaluate the generated results of LLMs from multiple perspectives, including the hallucination rate in generated references, factual consistency, and semantic coverage compared to human-written content.
    
    \item Finally, we assess five LLMs using the proposed framework. By analyzing the experimental results, we find that hallucinated references remain a prevalent issue for current LLMs. Furthermore, the performance of LLMs in writing literature reviews varies across different disciplines. 
\end{itemize}

\section{Related Work}
Recent studies have explored LLMs for literature review generation. For example, \citet{Wang_Guo_Yao_Zhang_Zhang_Wu_Zhang_Dai_Zhang_Wen_et_2024} proposed AutoSurvey, which incorporates up-to-date papers via retrieval-augmented generation. \citet{Agarwal_Sahu_Puri_Laradji_Dvijotham_Stanley_Charlin_Pal_2024} examined zero-shot LLM review generation using a two-step retrieval and outlining process. More recently, \citet{Liang_Yang_Wang_Tang_Zheng_Song_Lin_Yang_Niu_Wang_etal_2025} presented SurveyX, an efficient system that optimizes retrieval, extraction, and outline generation, supporting multimodal outputs such as figures and tables.

Additionally, recent efforts to evaluate literature review generation by LLMs have increasingly focused on assessing hallucinations in reference citations. For instance, \citet{Chelli_Descamps_Lavoue_2024} analyzed hallucination rates in 11 systematic reviews on shoulder rotator cuff pathology generated by ChatGPT, GPT-4, and Bard, finding Bard exhibited significantly higher hallucination rates. Similarly, \citet{Agrawal_Suzgun_Mackey_Kalai} evaluated hallucinations across 200 computer science topics by generating reference titles with LLMs and verifying their existence via the BING Search API, further probing whether LLMs could detect hallucinated references through direct and indirect queries. \citet{Athaluri_Manthena_Kesapragada_Yarlagadda_Dave_Duddumpudi} examined hallucinations in 50 ChatGPT-generated research proposals, manually validating references and DOIs using Scopus, Google, and PubMed, reporting 109 valid DOIs among 178 references. Additionally, \citet{Aljamaan_Temsah_Altamimi_Al-Eyadhy_Jamal_Alhasan_Mesallam_Farahat_Malki_2024} introduced the Reference Hallucination Score (RHS) by generating references for five medical topics across multiple LLMs, assigning weighted hallucination scores to citation components such as title and publication date. While these studies provide valuable insights, they are generally limited to specific domains and rely heavily on manual evaluation, lacking a comprehensive, scalable assessment framework for LLM reference hallucinations.

\section{Methodology}

\begin{figure*}[h]
    \centering
    \includegraphics[width=10.6cm, height=4cm]{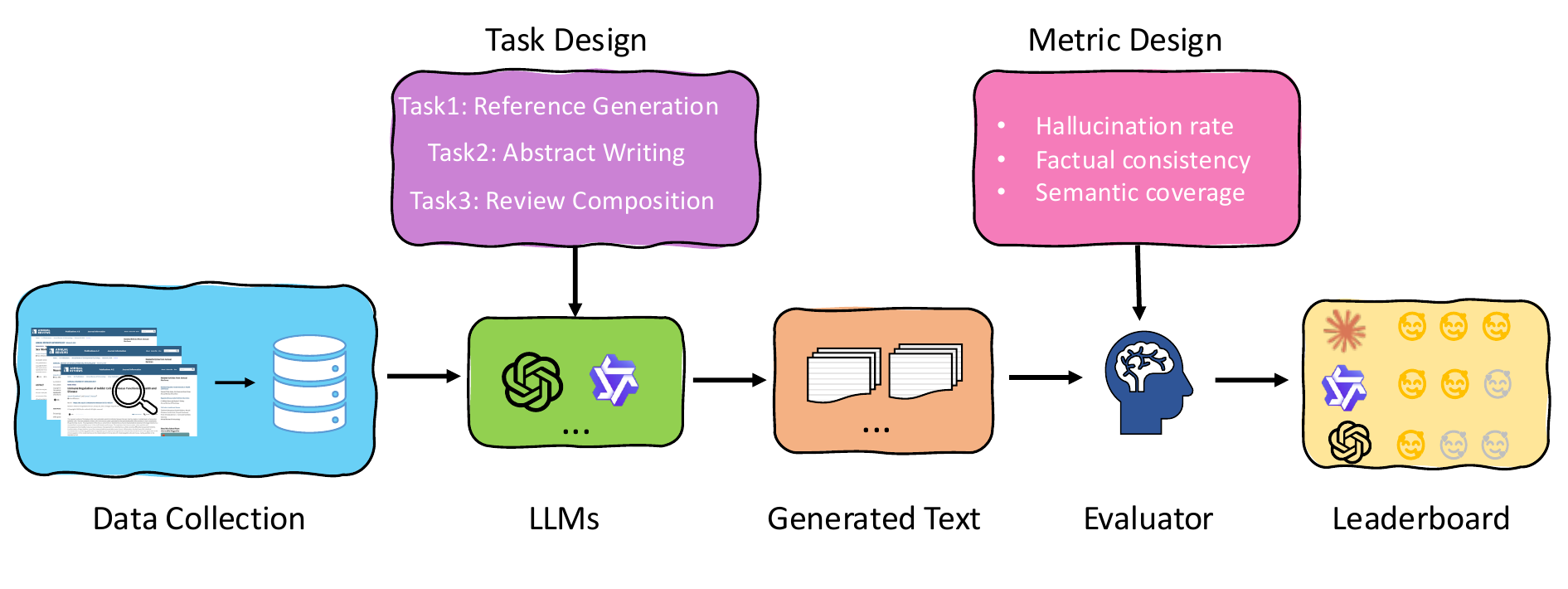}
    \vspace{-4mm}
    \caption{Illustration of the evaluation framework.}
    \label{f1}
    \vspace{-2mm}
\end{figure*}

In this section, we propose a framework for evaluating LLMs' literature review writing ability. The framework, as shown in Figure~\ref{f1}, consists of three main stages: dataset construction and task design for evaluation, collection LLM-generated output, and assessment of the generated output.

\subsection{Dataset Construction}

Assessing the ability to write literature reviews is a challenging task, as evaluating the quality of content is inherently complex. In this paper, we use human-written reviews as the gold standard, which simplifies the evaluation process to some extent. As illustrated in Figure \ref{f1}, we first collect publicly available information of literature reviews (i.e., the title, authors, abstract, and keywords) from the Annual Reviews website (\url{https://www.annualreviews.org/}). Annual Reviews, an independent nonprofit publisher, produces 51 review journals spanning various scientific disciplines. Invited experts write comprehensive, authoritative reviews that synthesize and summarize the most significant primary research literature in their field, providing a valuable resource for researchers to stay current with the latest developments. We crawl all articles published in 2023, including their title, keywords, abstracts, contents, and references, and then clean them to create the experimental dataset.

Then, the dataset $D$ is the article set from 51 journals, $D= \{p_0,...,p_i,...,p_M\}$, where $M$ represents the number of articles. Each article $p_i =\{t_i, w_i, a_i, c_i, R_i\}$, where $t_i$, $w_i$, $a_i$, $c_i$, $R_i$ represent the title, keywords, abstract, context, and reference set $R_i = \{r_1,...,r_k,...,r_K\}$, and $K$ represents the length of the reference set.

\subsection{Task Design}

Since literature review writing primarily involves the collection and synthesis of relevant research, we design three independent tasks as follows to evaluate LLMs' capabilities in different aspects of literature review writing.

\begin{itemize}
    \item \textbf{Reference Generation:} 
    Given the article title $t_i$ and keywords $w_i$, ask LLMs to find the $N$ most relevant studies $R_i^g = \{r_1^g,...,r_n^g,...,r_N^g\}$ to the research topic. Each citation study must include 7 metadata elements: title, authors, journal, year, volumes, first page, and last page, $r_n^g = \{T, A, J, Y, V, FP, LP\}$. In this task, we evaluate whether LLMs can recommend reliable references based on the given topic. \textit{Note that these references are not reused in later tasks.}

    \item \textbf{Abstract Writing:} Given an article title $t_i$ and its associated keywords $w_i$, the LLMs are prompted to generate an abstract $a_i^g$ that aligns with the research topic. The length of the generated abstract is constrained to match that of the original. This task serves as a proxy for literature review planning, as abstracts often outline the key components of a study—such as its objectives, methods, and covered subtopics—which are also critical in structuring comprehensive literature reviews. By evaluating the model’s ability to generate coherent and topic-relevant abstracts, we assess its potential to assist researchers in the early planning stages of literature review writing.

    \item \textbf{Review Composition:} Given the article $t_i$ and keywords $w_i$, and abstract $a_i$, ask LLMs to write a short literature review $c_i^g$ according to the research topic provided in the title, keywords, and abstract. To facilitate evaluation and accommodate computational budget constraints, the length of each literature review is limited to approximately 1000 words. LLMs also need to back up claims by citing relevent studies $R_i^g = \{r_1^g,...,r_n^g,...,r_N^g\}$ (with a total of $N$ citations in the literature review). These citations are newly generated to support the content of the review.
    In this task, we evaluate whether LLMs can write a high-quality literature review and cite truth studies.

\end{itemize}

Three task prompts are shown in Appendix Table~\ref{at1}.

\subsection{Evaluation Metrics}
\label{3.1}
Based on the type of generated text, we divide the evaluation of the model's results into two parts: first, the hallucination rate of the references generated by LLMs, and second, a comparison of the generated context with human-written results, including two dimensions: factual consistency and semantic coverage.

\textbf{Reference hallucination evaluation metrics.} Given that LLMs are trained on vast corpora, including academic sources, we aim to evaluate whether they can generate true references. In this section, we introduce the calculation process of the reference precision $Precison$, reference recall $Recall$, $F1$, and title search rate $S_t$ for each LLM. \textbf{A higher precision metric indicates a lower hallucination rate}. A higher recall indicates that the LLM-generated references cover more of the ground-truth citations used by human authors, reflecting a better ability to identify key prior work relevant to the topic. 

For each article $p_i \in D$, each LLM generates $N$ references $R_i^g = \{r_1^g,...,r_n^g,...,r_N^g\}$ in both Reference Generation and Review Composition tasks, each $r_n^g$ and includes 7 elements. Each element corresponding to a state label represents whether it is accurate or not $\{e_d\}_{d=0}^6$, $e_d$ = 1 or 0. 

Next, we describe how to obtain $\{e_d\}_{d=0}^6$. First, we use the generated titles $T$ and the first author in $A$ as the queries and search them separately from external academic search engines. This results in two sets of candidate articles, $Z_t$ and $Z_a$ respectively, We then merge the two sets and remove duplicates to obtain the final candidate set $Z =\{z_1,...,z_j,...,z_J\}$. Subsequently, we compare the generated $r_n^g$ with the article $z_j$ from candidate sets $Z$. For example, if the title of a candidate article $z_j$ matches the title of $r_n^g$, then $e_0 = 1$. Finally, we find the best candidate article based on the sum of $\{e_d\}_{d=0}^6$, and the one with the largest sum is the best candidate article $z_j$ of $r_n^g$. 

Then, we compare the alignment degree between the generated reference $r_n^g$ and the best-matching candidate article $z_j$ to determine whether $r_n^g$ corresponds to a real article (as shown in Eq.~\ref{e1}).
We consider $r_n^g$ to be reliable under either of the following two conditions:

\textbf{Title-based matching:} If the title $T$ is correct (i.e., $e_0 = 1$), and at least one other metadata element (e.g., author, journal, year, etc.) also matches, the reference is deemed reliable. To allow for minor variations, we consider the title to be correct if it achieves a match rate of at least 80\% with the ground-truth title—a threshold determined through human evaluation.

\textbf{Metadata-based matching:} If the title $T$ is incorrect (i.e., $e_0 = 0$), we still consider the reference reliable if at least three of the remaining metadata elements (author, journal, year, volume, first page, last page) match those of a real article. This allows us to identify true references even when the title is noisy or incomplete.
\begin{equation}
\small
\label{e1}
\text{True}(r_n^g) =
\begin{cases} 
1 & \text{if } \left( e_0 = 1 \text{ and } \sum_{i=1}^6 e_i \geq 1 \right) \\
  & \text{or } \left( e_0 = 0 \text{ and } \sum_{i=1}^6 e_i \geq 3 \right) \\
0 & \text{otherwise}
\end{cases}
\vspace{-2mm}
\end{equation}
For each paper $p_i$ in the dataset, we compute the reference precision of the LLM-generated references, denoted as $\textit{Precision}{(p_i)}$, as defined in Eq.~\ref{e2}.
We then obtain the overall Precision score for each LLM by averaging $\textit{Precision}{(p_i)}$ across all papers in the dataset, as shown in Eq.~\ref{e3}.
\begin{equation}
\small
Precision_{(p_i)}  = \frac{1}{N}\sum_{n=0}^{N} \text{True}(r_n^g)
\label{e2}
\vspace{-2mm}
\end{equation}
\begin{equation}
\small
Preicison  = \frac{1}{M}\sum_{i=0}^{M} Precison{(p_i)}
\label{e3}
\vspace{-1mm}
\end{equation}
Precision is measured by comparing the LLM-generated references with external academic databases. We also evaluate \textbf{Recall} by comparing the references generated by the LLM with those cited in the human-written original articles. The key difference between precision and recall in our setting lies in the candidate set $Z$: for precision, 
$Z$ is constructed from external academic search results, whereas for recall, 
$Z$ consists of the references actually cited in the human-written articles. Based on precision and recall, we further compute the \textbf{F1} score, which serves as a harmonic mean to reflect the overall accuracy of the reference generation.

Additionally, the title is intuitively the most critical element in determining the faithfulness of a generated reference. In the work of~\citet{Agrawal_Suzgun_Mackey_Kalai}, ground-truth labels were assigned based on results returned by the Bing Search API. Inspired by their approach, we also calculate the title search score for each LLM to estimate how many generated titles correspond to real publications.
\begin{equation}
\small
S_t = \frac{1}{MN} \sum_M \sum_N s_{p_i}^{(n)}
\label{e6}
\vspace{-2mm}
\end{equation} 
\begin{equation}
\small
\label{e7}
s_{p_i}^{(n)} = \sum_{r_n^g \in R_i^g} \begin{cases}
1 & \text{if the $T \in r_n^g$  has return value }\\
&\text{from external Scholar API,} \\
0 & \text{otherwise}
\end{cases}
\vspace{-1mm}
\end{equation}
Here, $s_{p_i}^{(n)}$ indicates whether the generated title in reference $r_n^g$ for paper $p_i$ can be found using an external academic search engine. This metric helps estimate the proportion of references with verifiable titles among the total generated references. 

\textbf{Context evaluation metrics.} In our study, we use the human-written article as the gold truth and then evaluate LLM-generated context from \textbf{factual consistency} and \textbf{semantic coverage} aspects. 
The resemblance of natural language inference (NLI) to factual consistency evaluation has led to utilizing NLI models for measuring factual consistency~\cite{Gao_Yen_Yu_Chen_2023}.
Encouraged by previous works, we also use the NLI method to evaluate the factual consistency between LLMs generated and human-written text. For example, we calculate the NLI score $Entail_{p_i}$ between the original article abstract $a_{i}$ and the LLM-generated abstract $a_{i}^g$ as follows.
\begin{equation}
\small
Entail_{p_i} = \theta_{\text{NLI}}(a_{i}^g, a_{i}) = 
\begin{cases}
1 & \text{if } a_{i}^g \text{ entails } a_{i} \text{,} \\
0 & \text{otherwise}
\end{cases}
\end{equation}
where $\theta_{NLI}$ denotes the NLI model. Finally, we obtain the NLI score $Entail$ for each model according to Eq~\ref{e9}.
\begin{equation}
    \small
    \label{e9}
    Entail = \frac{1}{M}\sum_{i=0}^{M}{Entail_{p_i}} 
    \vspace{-2mm}
\end{equation}
Additionally, we use commonly employed \textbf{semantic similarity metrics} and \textbf{Key Point Recall (KPR)} to calculate the semantic coverage between the context generated by LLMs and human-written context. Specifically, for the Abstract Writing task, we apply cosine similarity and the ROUGE metric for semantic coverage evaluation. 
For the Review Composition task, we use the ROUGE metric and KPR to measure the semantic coverage of the literature review generated by the LLMs relative to human-written content.

KPR, first proposed by ~\citet{Qi_Xu_Guo_Wang_Zhang_Xu_2024}, is a metric designed to evaluate the effectiveness of LLMs in utilizing RAG for long documents. Since human-written literature reviews are lengthy and difficult to compare directly, we adopt the KPR method to measure the extent to which LLM-generated content covers the key points in human-written literature reviews. Specifically, we first use GPT-4 to extract $q$ key points $X_i = [x_{i_1}, x_{i_2}, ..., x_{i_q}]$ from the human-written literature review $c_i$,
and then calculate the coverage of these key points by the model-generated literature review as Eq.~\ref{e10}.
\begin{equation}
    \small
    \label{e10}
    KPR = \frac{1}{M} \sum_{i=0}^{M} \frac{\sum_{x \in X_i} \theta_{\text{NLI}} (x, c_i^g)}{|X_i|}
    \vspace{-2mm}
\end{equation}
where $c_i^g$ denotes the literature review generated by LLMs.

Finally, we also concatenate key points and compute the ROUGE metric between the key points and $c_i^g$.

\section{Experiments}

\subsection{Experimental Settings}

\textbf{Dataset.} We collect 1,105 literature review articles published in 2023 from the Annual Reviews website. The distribution of articles across journals is shown in Appendix~\ref{a2}, Figure~\ref{f7}.

\textbf{LLMs Selection.} We evaluate five LLMs: \textbf{Claude-3.5-Sonnet-20240620}, \textbf{GPT-4o-2024-08-16}, \textbf{Qwen-2.5-72B-Instruct}, \textbf{DeepSeek-V3}, and \textbf{Llama-3.2-3B-Instruct}. All model outputs were generated via their official APIs with temperature set to 0 for consistency.

In Reference Generation and Review Composition tasks, we set $N$ as 10, each model generates 10 references.
For the generated reference evaluation, we use \textbf{Semantic Scholar} as the external database. Recently, LLMs-as-judges has become more common~\cite{Chen_Chen_Liu_Jiang_Wang_2024, Zheng_Chiang_Sheng_Zhuang_Wu_Zhuang_Lin_Li_Li_Xing_et_al_2023, Xing_Zhao_Wu_An_Chen_Li_Zhang_Dai_2024}. So, for the Abstract Writing task, we employ \textbf{TRUE}~\cite{honovich-etal-2022-true-evaluating}, along with \textbf{GPT-4o} as NLI models for evaluating factual consistency in context;
to compute semantic similarity, we use\textbf{ text-embedding-3-large} to convert texts into embeddings. For the Review Composition task, we employ \textbf{GPT-4o } as the NLI model in Eq~\ref{e10}, and set $q$ as 10.

\subsection{Main Results}

We present the results for the three tasks in Table~\ref{t1},~\ref{t2}, and~\ref{t3}. The performance of different models on each task is analyzed as follows.

\textbf{Results for Reference Generation.}
As shown in Table~\ref{t1}, Claude-3.5-Sonnet achieves the highest $F1$ score and $S_t$, while Llama-3.2-3B performs the worst on both metrics. Notably, Claude-3.5-Sonnet also achieves the highest precision and recall, indicating that it not only generates more correct references overall but also shares the highest overlap with those cited in the human-written article. When evaluating the author dimension of LLM-generated references, we consider the reference to match in this dimension if the first author is correctly matched. Applying this criterion results in a 1–3\% increase in F1 scores across all models. This suggests that generating complete and accurate author lists remains a major challenge for LLMs.

\begin{figure}
    \centering
    \includegraphics[width=1\linewidth]{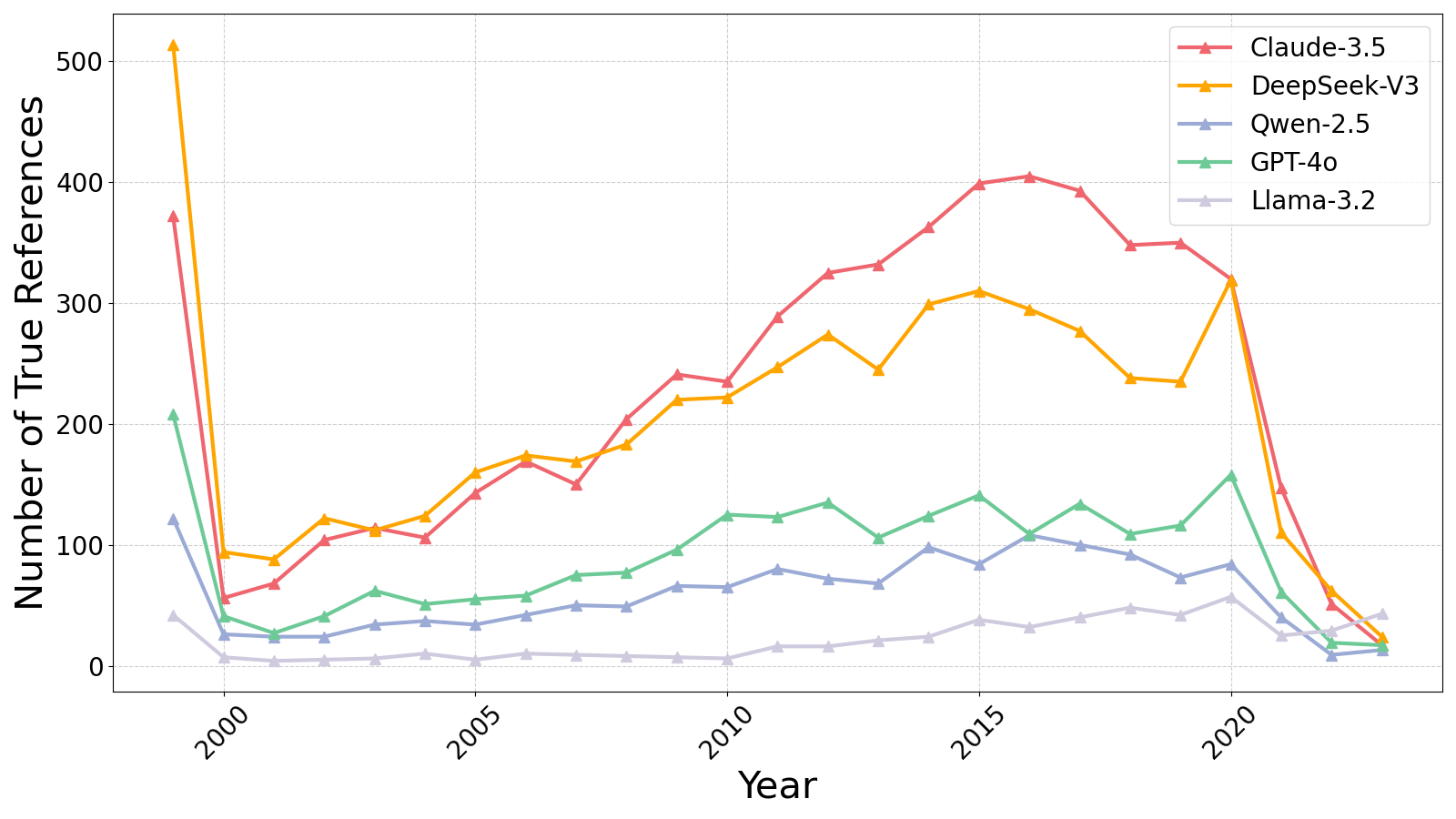}
     \vspace{-4mm}
    \caption{Distribution of LLM-generated true references over years.}
    \label{f1-1}
     \vspace{-4mm}
\end{figure}

We further conduct a year-wise analysis of the correctly generated references, as illustrated in Figure~\ref{f1-1}. The results reveal that the majority of accurate citations produced by the models are concentrated in the period between 2010 and 2020, a trend consistent across nearly all LLMs evaluated in this task.

\textbf{Results for Abstract Writing.} As shown in Table~\ref{t2}, Claude-3.5-Sonnet achieves the best overall performance across most evaluation metrics. It generates abstracts with the highest average semantic similarity to human-written ones (81.17\%) and shows strong factual consistency, achieving a TRUE score of 78.10\%. DeepSeek-V3 also performs well in factual consistency, with the highest GPT-4o-based assessment score (96.84\%). In contrast, Llama-3.2-3B obtains the highest ROUGE-L score but does not show clear advantages on other metrics. These results highlight the importance of using multiple evaluation metrics to comprehensively assess the diverse outputs of LLMs.

\textbf{Results for Review Composition.} As shown in Table~\ref{t3}, compared to the Reference Generation task, all LLMs demonstrate a significant increase in Precision when generating references within the Review Composition task. Prior research indicates that grounding generated text with real external citations can effectively reduce hallucination rates~\citep{Gao_Yen_Yu_Chen_2023}. Consistently, our experiments reveal that when LLMs generate references alongside the literature review, the accuracy of these references improves markedly. This suggests a mutual constraint between the generated references and the review text, leading to enhanced overall reliability.

On the other hand, Claude-3.5-Sonnet achieves the highest performance on the \textit{KPR} metric, indicating that its generated literature reviews recall the greatest number of claims from the human-written versions. Meanwhile, the literature reviews produced by DeepSeek-V3 excel on the ROUGE metrics, demonstrating stronger overlap with reference texts in terms of lexical similarity.

\begin{table*}[h!]
\small
\centering
\setlength{\tabcolsep}{0.5mm}
\begin{tabular}{c|c|ccc|ccc}
\hline
\Xhline{1.2pt}
\multirow{2}{*}{\textbf{Models} }& \multicolumn{7}{c}{\textbf{Reference Generation}}  \\
\cline{2-8}
 & \textbf{$\textbf{\textit{S}}_\textbf{\textit{t}}$$\uparrow$} & \textbf{\textit{P}} & \textbf{\textit{R}}& \textbf{\textit{F1}}&  \textbf{\textit{P}} & \textbf{\textit{R}}& \textbf{\textit{F1(first author)}} \\
\hline

\multirow{1}{*}{Qwen-2.5-72B} & 21.80 & 12.25 & 12.60& 12.42 & 17.58 &13.27 & 15.12\\
\multirow{1}{*}{Llama-3.2-3B}  & 16.62&3.45 &8.48& 4.90&  6.95 & 8.67&7.72 \\
\multirow{1}{*}{DeepSeek-V3} & 56.04 &46.33 &19.72 & 27.66&50.66 & 20.50 &29.19\\
\multirow{1}{*}{GPT-4o} &32.07&21.65 &18.76& 20.10& 24.65  & 19.50 & 21.77\\
\multirow{1}{*}{Claude-3.5-Sonnet } & \textbf{64.82} & \textbf{51.59}& \textbf{24.34 }& \textbf{33.08}&\textbf{55.77} &\textbf{25.21} & \textbf{34.72}\\
\hline
\Xhline{1.2pt}
\end{tabular}
\caption{The experimental results of the five LLMs in Reference Generation. ``$S_t$'' refers to the title search rate as defined in Eq~\ref{e6}, while ``\textit{P}'' represents the \textit{Precision}, ``\textit{R}'' denotes the \textit{Recall}. ``first author'' refers to when evaluating the accuracy of references, the author dimension only comparing the first author.}
\label{t1}
 \vspace{-3mm}
\end{table*}

\begin{table*}[h!]
\centering
\small
\setlength{\tabcolsep}{0.5mm}
\begin{tabular}{c|c|cc|ccc}
    \hline
    \Xhline{1.2pt}
\multirow{2}{*}{\textbf{Models} }& \multicolumn{6}{c}{\textbf{Abstract Writing}}  \\
\cline{2-7}
& \textbf{Similarity$\uparrow$}  & \textbf{Entail(TRUE)$\uparrow$}  & \textbf{Entail(GPT-4o)$\uparrow$}  &  \textbf{ROUGE-1$\uparrow$} &  \textbf{ROUGE-2$\uparrow$}&  \textbf{ROUGE-L$\uparrow$}\\
\hline
\multirow{1}{*}{Qwen-2.5-72B}  & 80.22& 69.52 &95.02 & 40.61 &8.78 &20.12 \\
\multirow{1}{*}{Llama-3.2-3B} &
  79.28 & 62.39 & 92.14& 40.35 &8.96 &\textbf{20.52 }\\

\multirow{1}{*}{DeepSeek-V3} &
  80.96 & 78.55 & \textbf{96.84}& \textbf{41.13} &8.98 &20.33\\

\multirow{1}{*}{GPT-4o} & 80.96 &77.91&
96.50&40.70 &8.56
& 19.86\\

\multirow{1}{*}{Claude-3.5-Sonnet } &\textbf{81.17} & \textbf{78.90}&96.77&\textbf{41.13}&\textbf{8.99}
&20.00\\
\hline
\Xhline{1.2pt}
\end{tabular}
\caption{Compare the performance of four LLMs on Abstract Writing.}
\label{t2}
\end{table*}

\begin{table*}[h!]
\small
\centering
\setlength{\tabcolsep}{0.8mm}
\begin{tabular}{c|c|ccc|ccc|cccc}
    \hline
    \Xhline{1.2pt}
\multirow{3}{*}{\textbf{Models} }& \multicolumn{11}{c}{\textbf{Review Composition}}  \\
\cline{2-12}
 & \multicolumn{7}{c|}{\textbf{References}}&\multicolumn{4}{c}{\textbf{Literature Review}}   \\
\cline{2-12}
 & \textbf{$\textbf{\textit{S}}_\textbf{\textit{t}}$$\uparrow$}  & \textbf{\textit{P}} & \textbf{\textit{R}}& \textbf{\textit{F1}}&  \textbf{\textit{P}} & \textbf{\textit{R}}& \textbf{\textit{F1(first author)}}& \textbf{\textit{KPR$\uparrow$}} &  \textbf{ROUGE-1$\uparrow$} &  \textbf{ROUGE-2$\uparrow$}&  \textbf{ROUGE-L$\uparrow$}\\
\hline
\multirow{1}{*}{Qwen-2.5-72B} & 40.02 & 28.91& 17.36 & 21.69
&33.64&18.31 & 23.713 & 38.82 & 29.95& 9.01 &15.14\\
\multirow{1}{*}{Llama-3.2-3B}  &21.78& 4.86 & 8.28& 6.12& 7.28&8.44 &7.82  &29.07 & 28.07&7.77&15.46\\

\multirow{1}{*}{DeepSeek-V3}  & 62.29 &52.81 &26.79 & 35.55  & 55.38 & 27.30&  36.57 & 56.02 &\textbf{35.65}&\textbf{10.40}
&\textbf{17.46}\\

\multirow{1}{*}{GPT-4o} &60.05 & 50.62 &27.88&35.96& 54.16  & 28.86& 37.65& 59.18 & 30.78 &9.72 &15.54\\

\multirow{1}{*}{Claude-3.5-Sonnet } & \textbf{66.43}
&\textbf{59.06} & \textbf{31.90}&\textbf{41.42} &\textbf{63.06}  &\textbf{33.25} &\textbf{43.54}&\textbf{62.32} &28.59 &8.90 &14.41\\
\hline
\Xhline{1.2pt}
\end{tabular}
\caption{The experimental results of the four LLMs in Review Composition. ``$S_t$'' refers to the title retrieval rate as defined in Eq~\ref{e6}. While ``\textit{P}'' represents the \textit{Precision}, ``\textit{R}'' denotes the \textit{Recall}. ``\textit{KPR}'' means the 
\textit{Key Point Recall} rate.``first author'' refers to when evaluating the accuracy of references, the author dimension only comparing the first author. }
\label{t3}
\end{table*}

\begin{table}[h]
\small
\centering
\setlength{\tabcolsep}{0.5mm}
\begin{tabular}{c|c|c|c|c}
\Xhline{1.2pt}
\hline
\multirow{2}{*}{\textbf{Discipline}} & 
\multicolumn{2}{c|}{\textbf{Citation Count}} & 
\multicolumn{2}{c}{\textbf{Precision}} \\
\cline{2-5}
& {DeepSeek} & {Claude} & {DeepSeek} & {Claude} \\
\hline
Biology      & 763  & 678  & 55.55 & 58.00 \\
Mathematics  & \textbf{2288} & \textbf{1984} & \textbf{60.00} & \textbf{62.22} \\
Physics      & 894  & 652  & 47.62 & 56.19 \\
Chemistry    & 1334 & 1079 & 43.14 & 43.80 \\
Social Science & 1321 & 1151 & 46.80 & 56.70 \\
Technology   & 904  & 748  & 44.88 & 49.01 \\
\hline
\Xhline{1.2pt}
\end{tabular}
\caption{Average citation counts and reference precision across disciplines.}
\label{t4}
\vspace{-6mm}
\end{table}


\begin{figure}[htp]
    \centering
    \subfigure[Reference Generation] {
     \label{f2a}     
    \includegraphics[width=0.45\columnwidth]{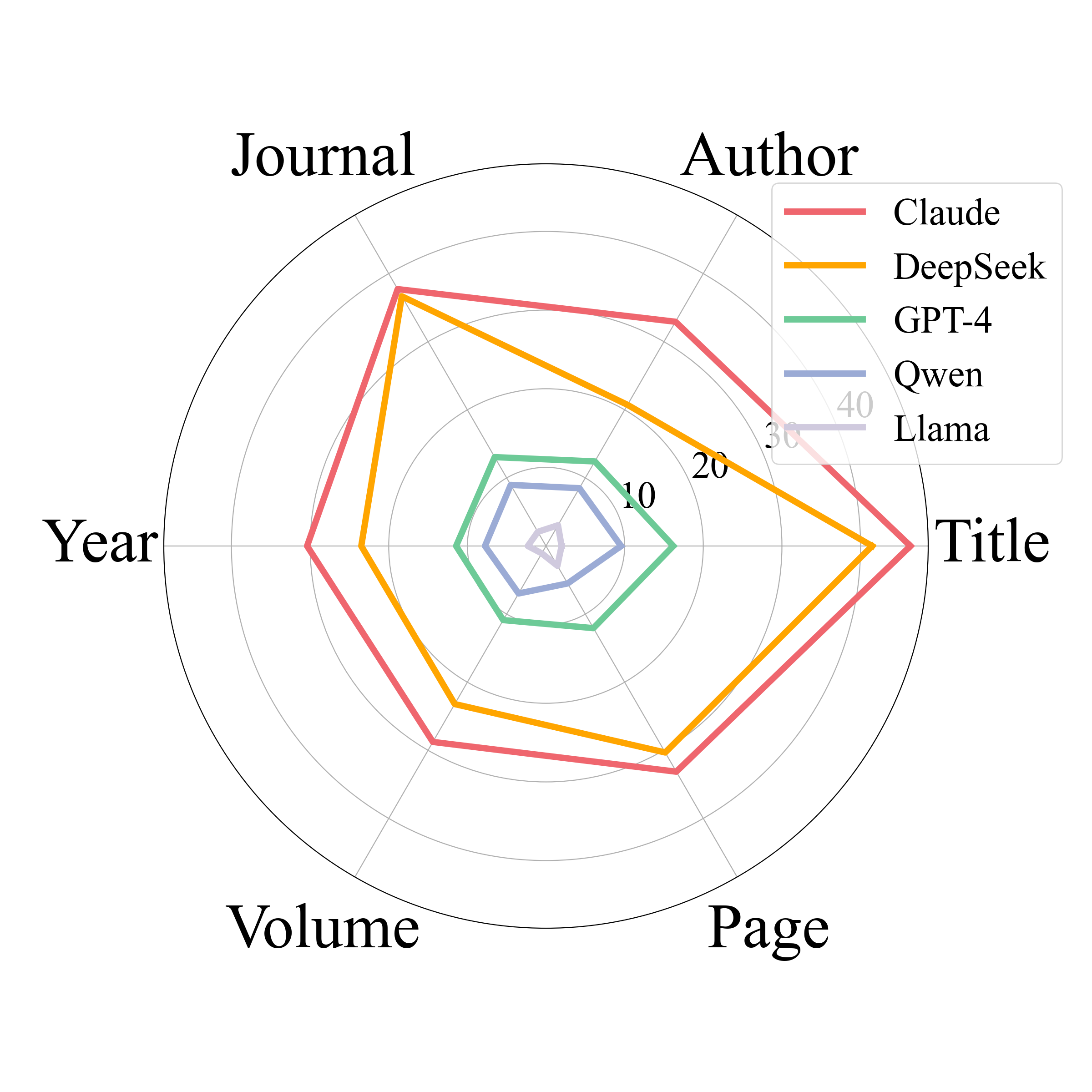}  
    } 
    \subfigure[Review Composition] {
    \label{f2b}     
    \includegraphics[width=0.45\columnwidth]{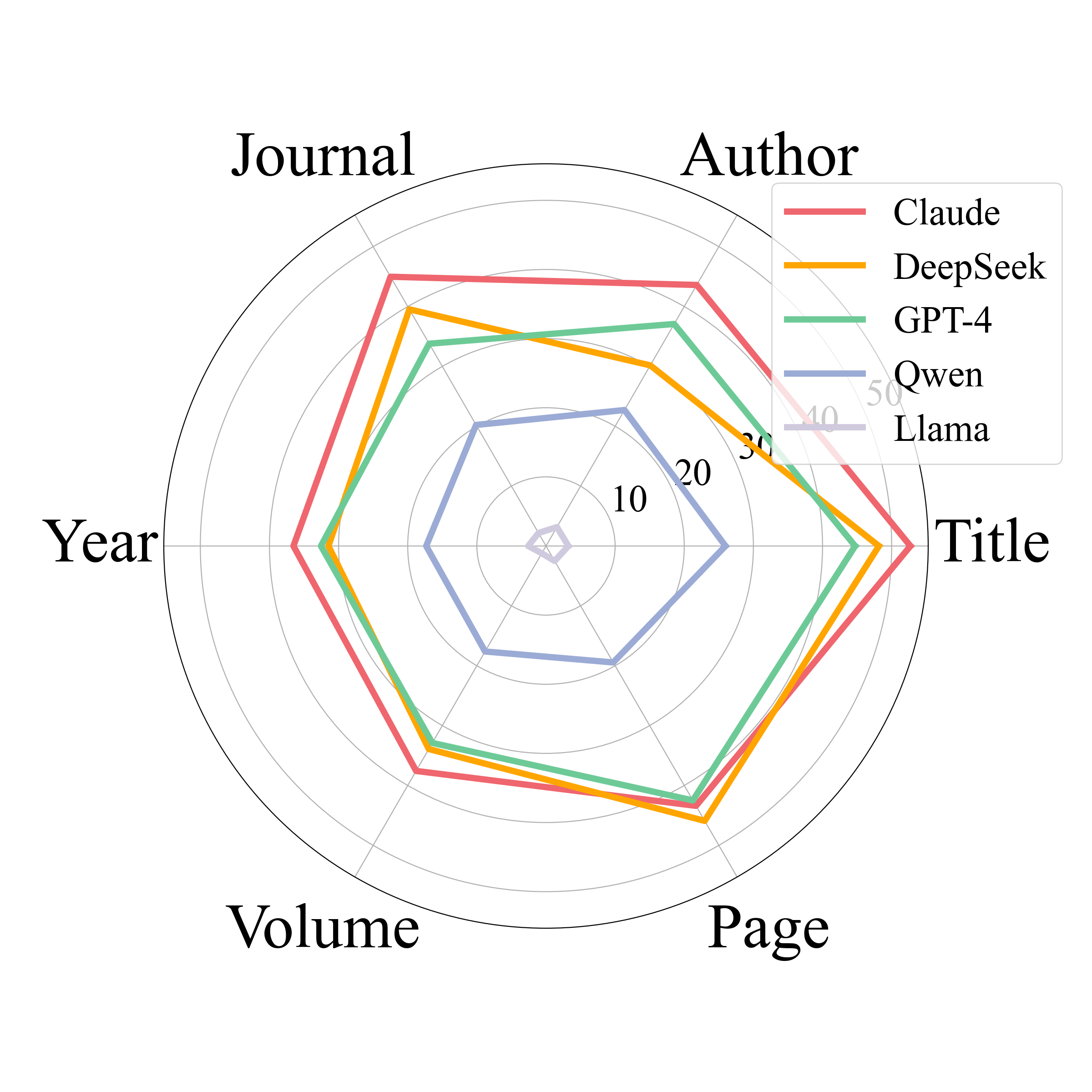}  } 
    \caption{Radar chart of the accuracy of LLM-generated references across various dimensions.}
    \label{f2}
     \vspace{-4mm}
\end{figure}




\begin{figure*}[htp]
    \centering
    \subfigure[Reference Generation: Precision $\uparrow$]{
     \label{f4a}     
    \includegraphics[width=7cm, height=4cm]{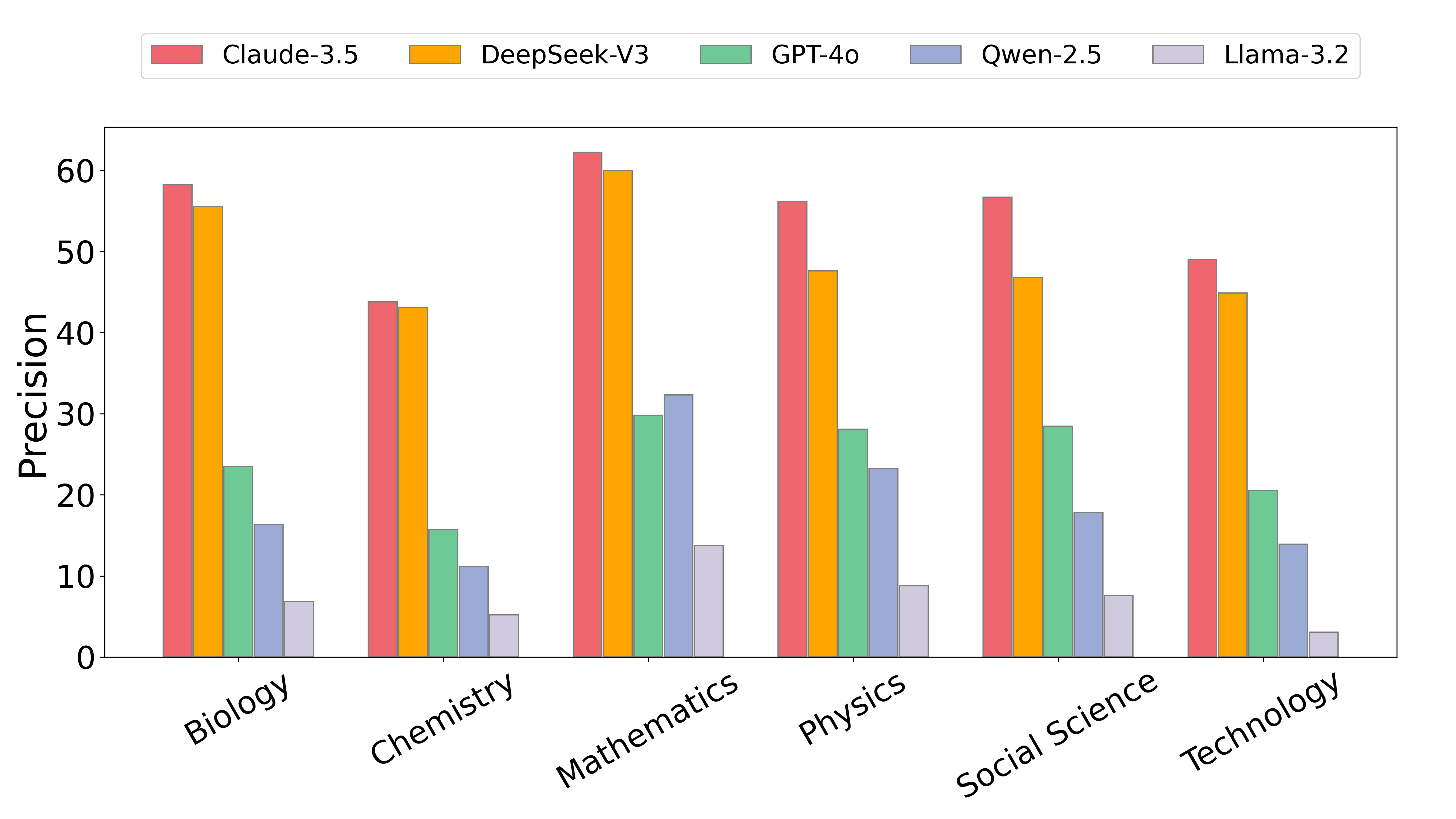}  
    } 
    \subfigure[Abstract Writing: NLI scores (TRUE)$\uparrow$]{
     \label{f4b} 
    \includegraphics[width=7cm, height=4cm]{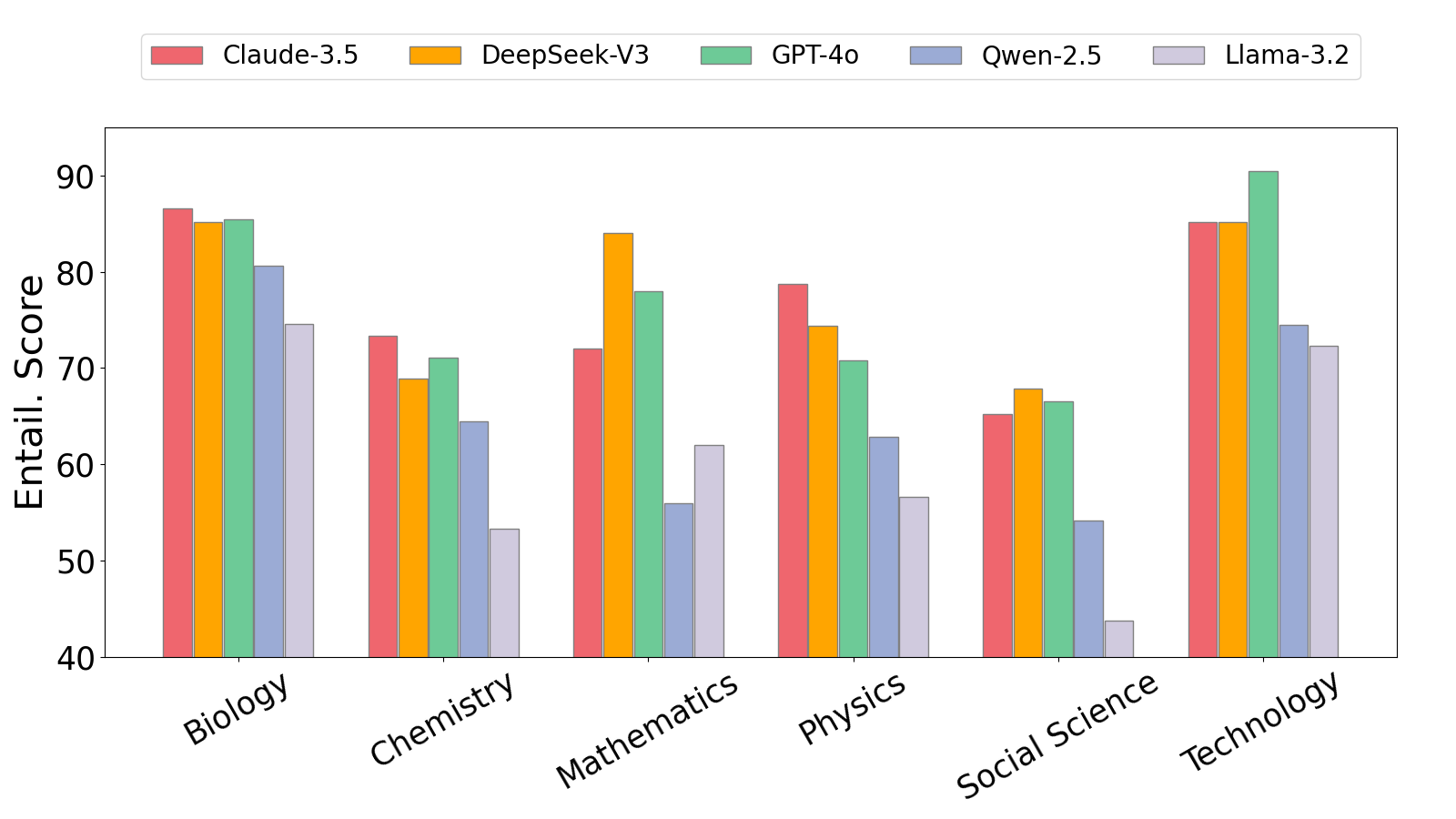}
    }  
    
     \vspace{-2mm}
    \subfigure[Review Composition: Precision $\uparrow$]{
     \label{f4c}     
    \includegraphics[width=7cm, height=4cm]{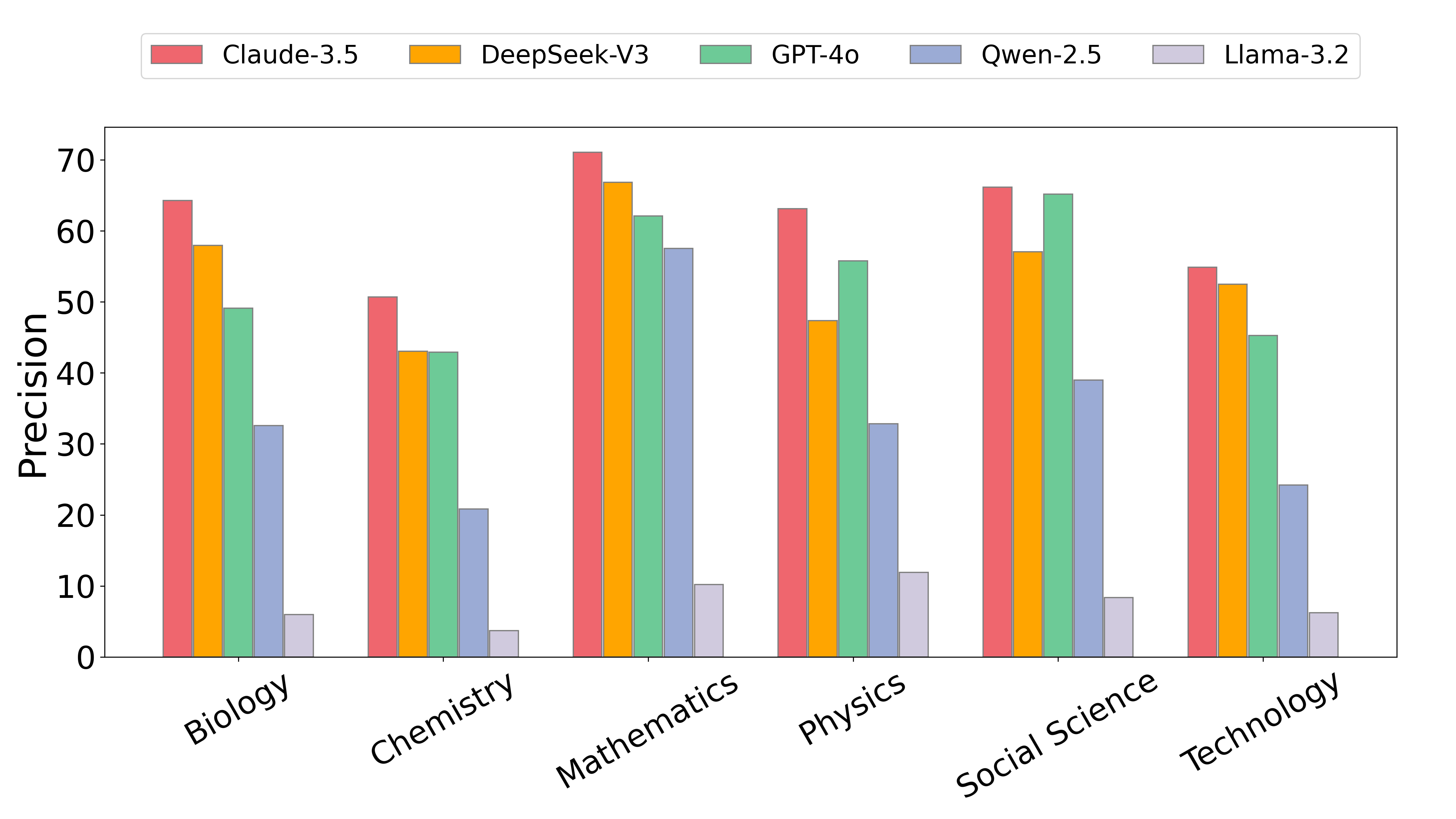}  
    } 
    \subfigure[Review Composition: KPR scores $\uparrow$]{
    \label{f4d}     
    \includegraphics[width=7cm, height=4cm]{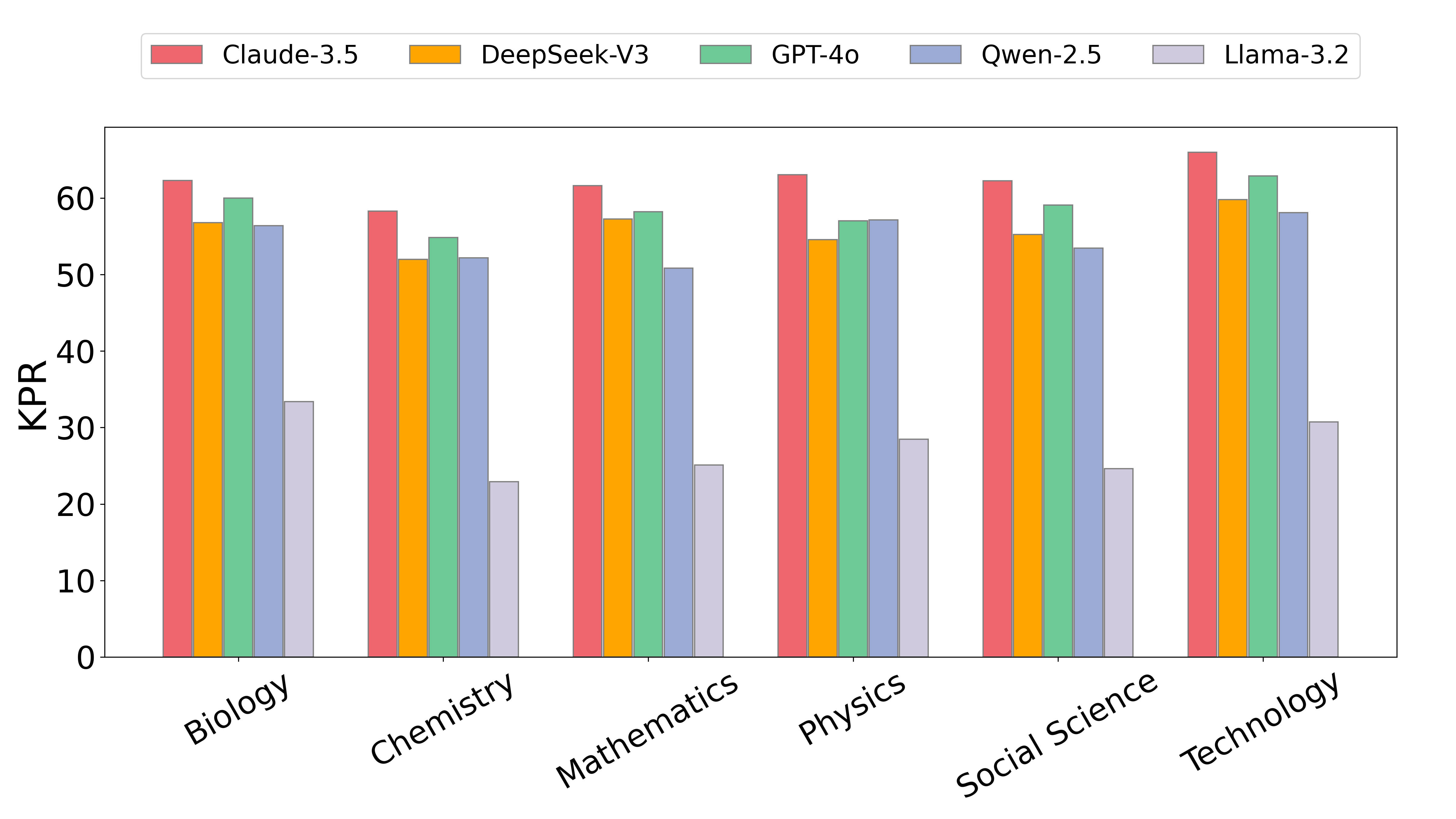}} 
     \vspace{-2mm}
    \caption{Three tasks evaluation scores across different disciplines.}
    \label{f4} 

\end{figure*}

\subsection{Analyze LLM-Generated References from Different Dimensions}

In both Reference Generation and Review Composition tasks, we ask LLMs to generate references. The overall performance was discussed in the previous section. In this section, we provide a detailed comparison of the accuracy of LLM-generated references across various dimensions, as shown in Figure~\ref{f2}. As seen in Figure~\ref{f2a}, Claude-3.5-Sonnet demonstrates a clear advantage over other models across all dimensions in the Reference Generation task. Additionally, the accuracy of reference generation in the Reference Generation task for Claude-3.5-Sonnet, GPT-4o, and Qwen-2.5-72B follows a consistent trend across all dimensions, with the highest accuracy observed in the title dimension. Accuracy for journal name, page, and author is also relatively high. However, DeepSeek-V3 performs worse in the author dimension compared to the other dimensions. In contrast, Llama-3.2 demonstrates higher accuracy in the page and author dimensions than in other dimensions. However, overall, Llama-3.2-3B does not exhibit a competitive advantage in reference generation accuracy.

Next, we examine the accuracy of LLM-generated references across various dimensions in Review Composition, as shown in Figure~\ref{f2b}, and comparing it with Figure~\ref{f2a}, we observe improvements across all dimensions for Claude-3.5-Sonnet, DeepSeek-V3, GPT-4o, and Qwen-2.5-72B, with particularly obvious gains in the author dimension. The possible reason is that, in the generated text, the LLMs tend to cite the first author's name, which may lead the models to place more emphasis on this dimension. Notably, the accuracy of DeepSeek-V3 and GPT-4o in certain dimensions approaches or even exceeds that of Claude-3.5-Sonnet. However, the performance of LLaMA-3.2-3B remains suboptimal.

\subsection{Cross-Disciplinary Analysis}

In this section, we compare the performance of LLMs across different disciplines. First, based on Dewey's Decimal Classification, we categorize 51 journals into five disciplines: Biology, Chemistry, Mathematics, Physics, Social Science, and Technology. After categorization, there are 460 articles in the Biology category, 90 in Chemistry, 50 in Mathematics, 113 in Physics, 299 in Social Science, and 94 in Technology.


We then present bar charts in Figures ~\ref{f4}, which illustrate the performance of different models across various tasks and disciplines. 

First, we observe that in the Reference Generation task, as shown in Figure~\ref{f4a}, almost all models exhibit the highest precision in the Mathematics discipline and the lowest precision in the Chemistry discipline. 
To validate these differences, we conduct one-way ANOVA tests for each LLM across five disciplines. Significant differences are found for all models except Llama-3.2-3B. Detailed ANOVA results are reported in Appendix~\ref{appendix:anova}.


Secondly, as shown in Figure~\ref{f4b}, the NLI scores evaluated by TRUE in the Abstract Writing task indicate that all models perform the worst in Social Science. GPT-4o performs best in Technology, while Claude 3.5-Sonnet achieves the highest performance in Biology.
One-way ANOVA tests reveal significant differences across disciplines for all models. See Appendix~\ref{appendix:anova} for detailed results.

Thirdly, we examine the references precision of each model across five disciplines in the Review Composition task, as illustrated in Figure~\ref{f4c}. 
It is evident that the precision of Claude 3.5 Sonnet, DeepSeek-V3, and GPT-4o is significantly higher than that of Qwen-2.5-72B and LLaMA-3.2-3B across all disciplines. Furthermore, Claude 3.5 Sonnet, DeepSeek-V3, and Qwen-2.5-72B exhibit the highest precision in Mathematics, while GPT-4o performs best in Social Science.
ANOVA tests confirm significant differences across disciplines for all models (see Appendix~\ref{appendix:anova}).
 

Finally, we observe the KPR metric across different disciplines in Review Composition, as shown in Figure~\ref{f4d}. The results from the figure indicate that the differences between models—Claude-3.5-Sonnet, DeepSeek-V3, GPT-4o, and Qwen-2.5-72B—are not significant across various disciplines, a finding that is also supported by statistical tests (see Appendix~\ref{appendix:anova}).


\textbf{Citation Frequency and Precision Across Disciplines.} Additionally, we report statistics on the citation frequency of correctly generated references by LLMs in the Reference Generation task, as shown in Table~\ref{t4}, using Claude-3.5 and DeepSeek-V3 as examples. The data indicates that the references generated by the LLMs are highly cited, which might be due to their frequent presence in online sources, making them more likely to appear in the LLMs training datasets. As a result, LLMs tend to generate more accurate metadata (e.g., author, year) for these well-known references.

Furthermore, when analyzing different disciplines, we observe that the Mathematics discipline has the highest precision, and the relevant references generated for Mathematics also have the highest citation count. We compute the correlation between citation precision and average citation counts, finding that the correlation coefficient for Claude-3.5 is 0.4, and for DeepSeek-V3 it is 0.51, indicating a positive relationship between the two.

\subsection{Human Evaluation}
 
To evaluate the reliability of our automatic assessment method for identifying hallucinated references, we conduct a comparative analysis involving 100 LLM-generated references. These references were assessed by three annotators and the final manual results were obtained by majority vote.
The results demonstrated a kappa agreement of 0.71 between the automatic and human assessments, signifying a relatively high level of consistency and supporting the reliability of our method. Furthermore, when using human assessment results as the gold standard, the automatic assessment method achieved an accuracy of 86\%, further validating its effectiveness.

\section{Conclusion}
In this paper, we present a framework to assess the literature review writing abilities of LLMs. This framework includes three tasks designed to evaluate LLMs' literature review writing capabilities. The generated outputs are then evaluated from multiple dimensions using various tools, such as Semantic Scholar and NLI models, focusing on aspects like hallucination rate, semantic coverage, and factual consistency compared to human-written texts. Finally, we analyze the performance of LLMs in writing literature reviews from the perspective of different academic disciplines.

\section*{Limitations}

In this paper, we evaluate the ability of LLMs to write literature reviews. However, several limitations remain:

First, instead of evaluating the generated reviews from conventional perspectives such as fluency or topic coverage, we primarily compare LLM-generated results with human-written ones. As such, our current evaluation metrics may not be comprehensive. In the future, we plan to incorporate additional aspects of review quality to improve the completeness of our evaluation. These may include the coverage of cited works (i.e., whether the review offers a comprehensive overview of the relevant field) and the coherence of the overall structure (i.e., whether the review is organized in a way that facilitates information-seeking).

Second, there is a possibility that our test data overlaps with the training data of the LLMs. When we initiated this study in August 2024, the dataset from the Annual Reviews website had not yet been updated to include 2024 articles, so we relied on the complete 2023 dataset. To mitigate potential data leakage, we plan to deploy a leaderboard on Hugging Face to continuously evaluate the performance of various LLMs in literature review writing, with real-time updates to the test dataset. However, due to the rapid iteration of LLMs, data leakage cannot be completely ruled out. That said, our experimental results—particularly those related to reference generation—show that all models still perform poorly. If data contamination were present, the actual scores would likely be lower than those reported. This reinforces, rather than undermines, our conclusion that significant challenges remain in using LLMs for literature review generation.

Additionally, when processing LLM-generated outputs, we frequently encountered abbreviated author names and journal titles. Although we have taken care to address these issues thoroughly (see Appendix~\ref{data_process}), minor discrepancies may still exist.

Finally, to verify the precision of LLM-generated references, we primarily used Semantic Scholar as our auxiliary tool. Although we also experimented with Google Scholar, its lack of an accessible API led us to rely on the freely available Semantic Scholar API for consistency and ease of access. However, this may have resulted in incomplete reference retrieval.

\section*{Ethics Statement}
The human evaluations conducted in this study were carried out by members of the research team. No personal or sensitive information was collected, and all participants were fully informed of the purpose of the evaluation. Therefore, the study does not raise any ethical concerns.

\bibliography{custom}
\bibliographystyle{acl_natbib}

\clearpage
\appendix
\onecolumn
\section{Prompts for Tasks}

\label{sec:appendix}

\begin{longtable}{|p{2.5cm}|p{12cm}|} 
    \caption{Prompts for tasks.} \label{at1} \\

    \hline
    \textbf{Prompt} & \textbf{Content} \\ \hline
    \endfirsthead

    \hline
    \textbf{Prompt} & \textbf{Content} \\ \hline
    \endhead

    \hline
    \endfoot

    \hline
    \endlastfoot
Prompt 1 & Imagine you are an experienced academic researcher with access to a vast library of scientific literature. I would like you to find the 10 studies that are most relevant to the research topic provided in the "Title" and the "Keywords" below.\newline
Please cite the studies according to the following JSON format. There is no need to provide any explanation before or after the JSON output. Ensure that the "authors" field lists the names of all authors and not exceeding 10 authors, and that there are no duplicate author names nor abbreviations such as "et al.".
\{
  "References": [
    \{
      "title": "",
      "authors": "",
      "journal": "",
      "year": "",
      "volumes": "",
      "first page": "",
      "last page": "",
   \}
  ]
\}\newline
Title: \textcolor{blue}{title}\newline
Keywords: \textcolor{blue}{keywords} \\ \hline

Prompt 2 & Imagine you are an experienced academic researcher with access to a vast library of scientific literature. I would like you to write an abstract according to the research topic provided in the "Title" and the "Keywords" below.
Please write the abstract for about \textcolor{blue}{xx} words, according to the JSON format as follows. There is no need to provide any explanation before or after the JSON output.
\{"Abstract": ""\}\newline
Title: \textcolor{blue}{title}\newline
Keywords: \textcolor{blue}{keywords} \\ \hline

Prompt 3 & Imagine you are an experienced academic researcher with access to a vast library of scientific literature. I would like you to write a literature review according to the research topic provided in the "Title", “Abstract” and "Keywords" below.

The literature review should be about 1000 words long. I would like you to back up claims by citing previous studies (with a total of 10 citations in the literature review). The output should be in JSON format as follows:
\{
  "Literature Review": "xxx",
  "References":  [
    \{
      "title": "",
      "authors": "",
      "journal": "",
      "year": "",
      "volumes": "",
      "first page": "",
      "last page": "",
    \}
  ]
\}\newline
The "Literature Review" field should be about 1000 words. The "References" field is a list of 10 references, and ensures that the "authors" field lists the names of all authors and not exceeding 10 authors, and that there are no duplicate author names nor abbreviations such as "et al.".\newline
Title: \textcolor{blue}{title}\newline
Keywords: \textcolor{blue}{keywords}\newline
Abstract: \textcolor{blue}{abstract} \\
\end{longtable}

\begin{figure}[H]
    \centering
    \includegraphics[width=12cm, height=4cm]{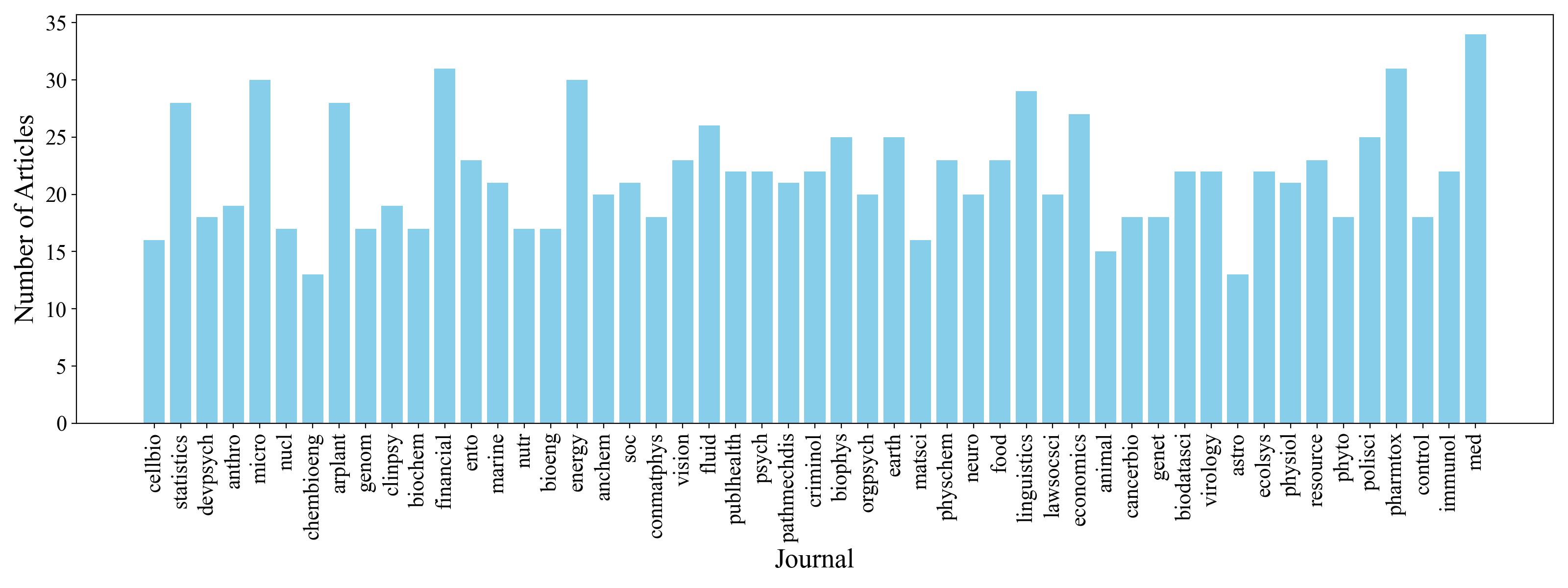}
    \caption{Statistics of dataset.}
    \label{f7}
\end{figure}

\twocolumn
\section{Data Distribution}
Statistics of the dataset are shown in Figure~\ref{f7}.
\label{a2}

\section{Comparison of LLM-Cited and Human-cited References from Different Dimensions}
\label{a3}
\begin{figure}[H]
    \centering
    \subfigure[Task1] {
     \label{f3a}     
    \includegraphics[width=0.45\columnwidth]{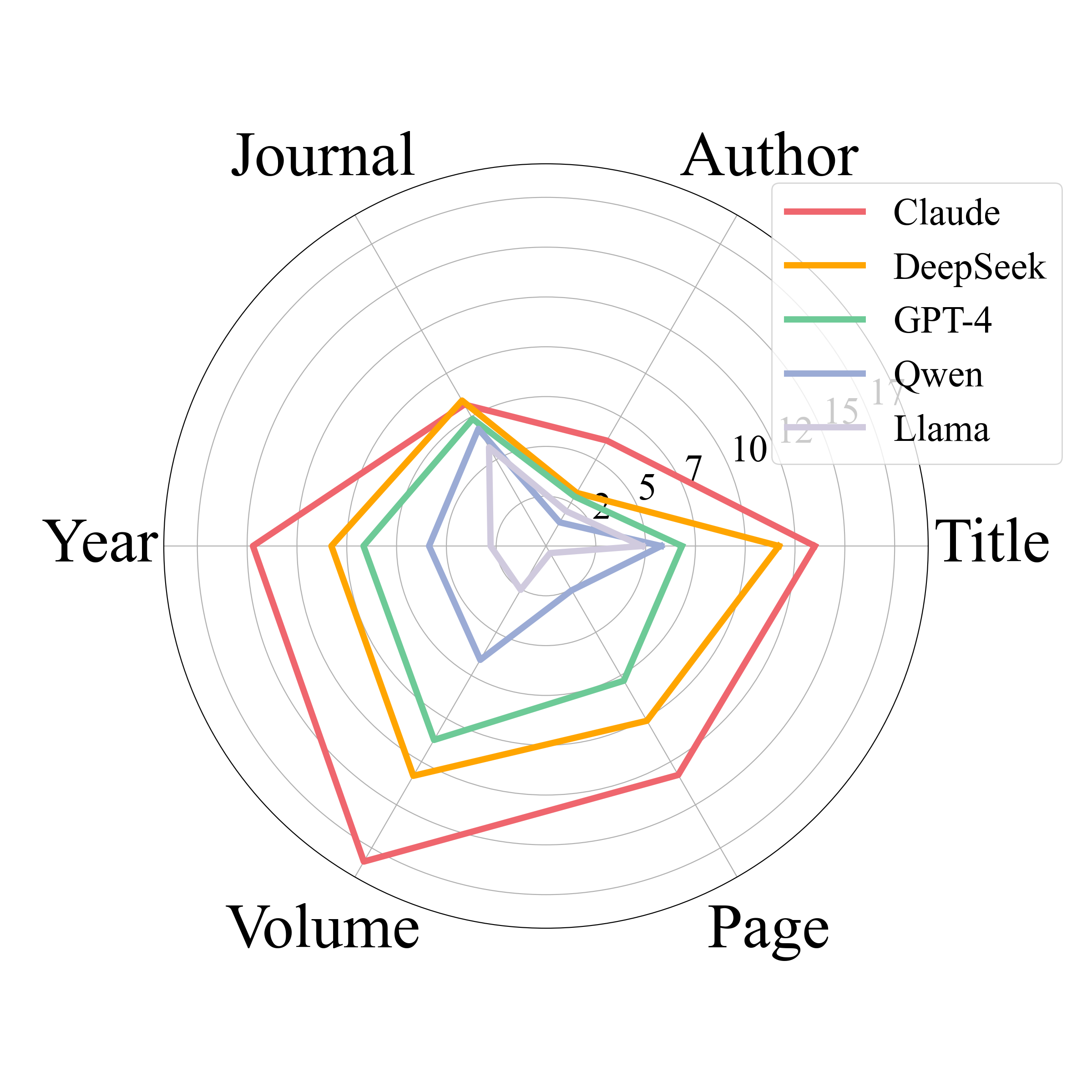}
    }
    \subfigure[Task3] {
    \label{f3b}     
    \includegraphics[width=0.45\columnwidth]{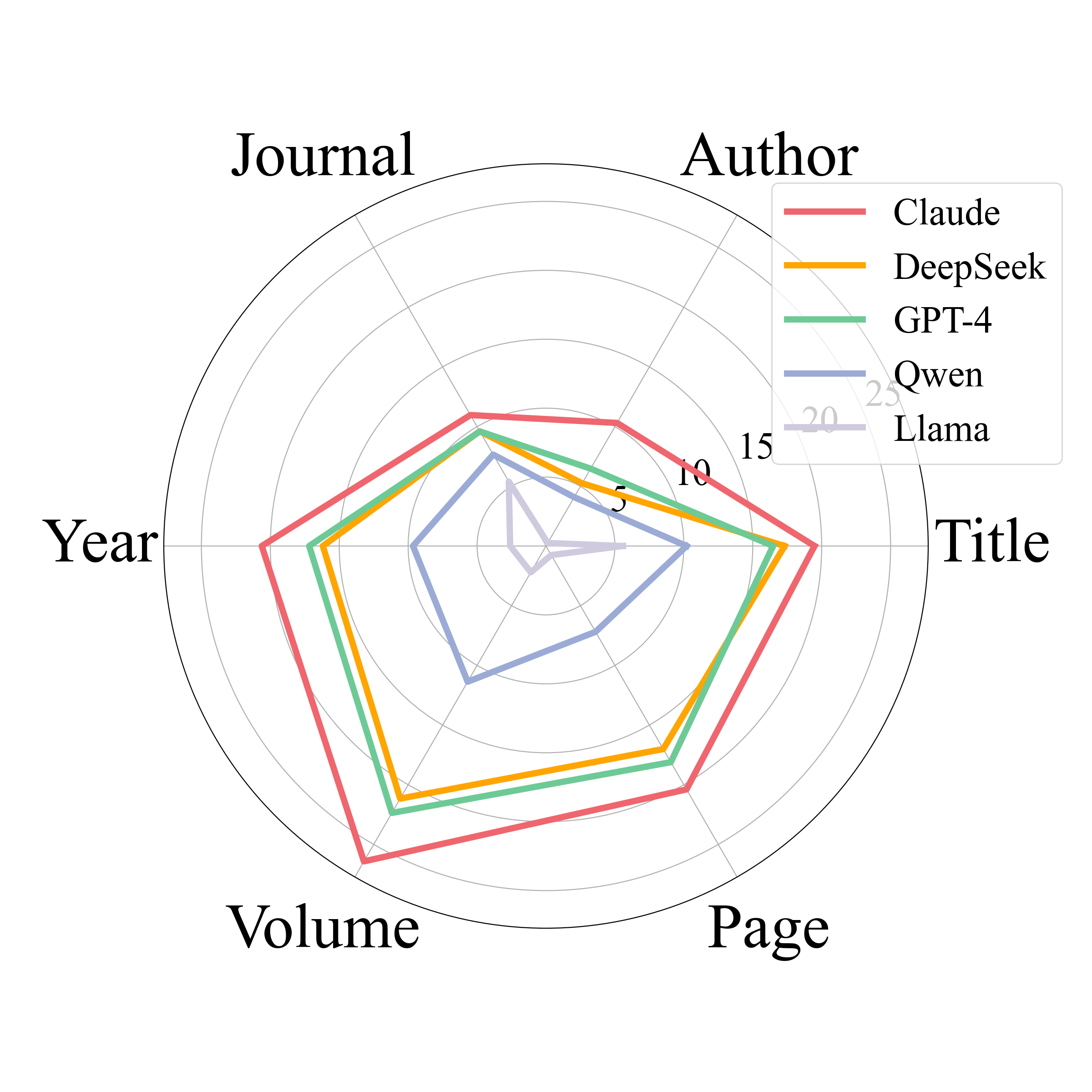}  
    }
    \caption{Radar chart of the accuracy of LLM-generated references with human-written references in the original article.}
    \label{f3} 
\end{figure}

We provide a more detailed comparison of the LLM-cited and human-cited references across various dimensions. As shown in Figure~\ref{f3}, for Reference Generation, we observe that the overlap rate is higher in the ``Title'' and other numerical dimensions, while the overlap rates for the ``Journal'' and ``Author'' dimensions are relatively lower.
For Review Composition, Claude-3.5-Sonnet and GPT-4o exhibit a higher overlap rate on the ``Author'' dimension compared to Reference Generation. This trend is consistent with the findings in Figure~\ref{f3}, as the citation of author names in the literature for Review Composition leads to the generation of more accurate author information.

\section{Discussion}
We select four LLMs for task evaluation and find that Claude-3.5-Sonnet outperforms DeepSeek-V3, GPT-4o, Qwen-2.5-72B, and Llama-3.2-3B across all three tasks, particularly excelling in the task of generating accurate references. This advantage is likely influenced by the training data of each model. Additionally, we observed that each model has different strengths across disciplines. Overall, for the reference generation task, nearly all models perform better in Mathematics, while their performance is weaker in Chemistry and Technology. However, when writing abstracts, all models exhibit the lowest factual consistency in Social Science, as indicated by the entailment scores, compared to human-written texts.

When comparing the references generated by the models in Reference Generation and Review Composition, we find that in Review Composition, nearly all models generate more accurate references. This suggests that LLMs cite references during the writing process, which improves the authenticity of the references. Moreover, the inclusion of the first author's name in the generated context also enhances the accuracy of the author dimension.

In the future, we will deploy this evaluation framework on Hugging Face to enable real-time assessments of newly released LLMs.

\section{Data Processing Strategy}
\label{data_process}

Author name and journal title variations often pose challenges when aligning LLM-generated references with articles from Semantic Scholar. To address this, we adopt the following normalization strategies:

Author names. When comparing author names between LLM-generated references and candidate articles, if an exact match is not found (e.g., ``John Smith''), we consider common variants such as ``Smith, John'', ``Smith, J.'', or ``J. Smith'' to account for different citation formats.

Journal titles. For journal names like Journal of Chemical Physics, we incorporate standard abbreviation forms (e.g., J. Chem. Phys.) based on widely used abbreviation conventions. Nonetheless, certain non-standard or ambiguous cases may still be unmatched.

\section{LLM Output Example}

Here is one example of the output generated by Claude-3.5-Sonnet for the literature review composition task.

\textit{\{"Literature Review": "CRISPR-Cas systems have emerged as a powerful tool in the field of biosensing, offering unprecedented capabilities for detecting nucleic acids, proteins, and small molecules. This review explores the technical developments and application prospects of CRISPR-based biosensing strategies, highlighting their potential to revolutionize diagnostics and molecular detection.The core biochemical properties of CRISPR-Cas systems provide the foundation for their application in biosensing. These properties include diverse reaction temperatures, programmability in design, high reaction efficiency, and recognition specificity. Chen et al. (2018) demonstrated the versatility of CRISPR-Cas9 in detecting DNA targets across a wide temperature range, from 20°C to 60°C, enabling its use in various environmental conditions. The programmability of CRISPR systems allows for the easy design of guide RNAs (gRNAs) to target specific sequences, as shown by Gootenberg et al. (2017) in their development of the SHERLOCK (Specific High-sensitivity Enzymatic Reporter unLOCKing) platform for nucleic acid detection.The high reaction efficiency of CRISPR-Cas systems contributes to their sensitivity in biosensing applications. Li et al. (2019) reported a CRISPR-Cas12a-based assay capable of detecting attomolar concentrations of DNA targets, demonstrating the potential for ultra-sensitive detection. The recognition specificity of CRISPR systems allows for precise discrimination between closely related sequences, as evidenced by the work of Myhrvold et al. (2018), who developed a CRISPR-based diagnostic platform capable of distinguishing between Zika and Dengue virus strains with single-base resolution.Recent efforts have focused on improving these core properties to enhance the performance of CRISPR-based biosensors. For example, Harrington et al. (2018) engineered Cas12a variants with improved activity and specificity, expanding the range of targetable sequences and reducing off-target effects. Similarly, Gootenberg et al. (2018) developed an enhanced version of the SHERLOCK system, incorporating lateral flow readouts and multiplexing capabilities, further improving its utility in point-of-care diagnostics.One of the key areas of technical development in CRISPR-based biosensing is improving sensitivity and quantification capabilities. Wang et al. (2020) introduced a CRISPR-Cas12a-based digital PCR method that achieved attomolar sensitivity and precise quantification of target nucleic acids. This approach combines the specificity of CRISPR recognition with the quantitative power of digital PCR, opening new avenues for highly sensitive and accurate molecular diagnostics.Multiplex detection is another important aspect of CRISPR biosensing development. Ackerman et al. (2020) demonstrated a CRISPR-Cas13-based multiplexed assay capable of simultaneously detecting and distinguishing multiple SARS-CoV-2 variants, showcasing the potential of CRISPR technology in rapid and accurate pathogen identification. The ability to detect multiple targets in a single reaction not only saves time and resources but also provides more comprehensive diagnostic information.Efforts to develop convenient one-pot assays have led to simplified and more user-friendly CRISPR-based detection methods. Joung et al. (2020) reported a one-pot, CRISPR-Cas12-based assay for SARS-CoV-2 detection that could be performed at a single temperature, eliminating the need for complex thermal cycling equipment. This advancement brings CRISPR-based diagnostics closer to point-of-care applications, where simplicity and ease of use are crucial.The creation of advanced sensors has further expanded the capabilities of CRISPR-based biosensing. Bruch et al. (2019) developed a CRISPR-Cas13a-based electrochemical sensor for nucleic acid detection, demonstrating the potential for integrating CRISPR technology with electronic readout systems. This approach offers the possibility of developing portable, electronic CRISPR-based diagnostic devices for rapid and sensitive detection in resource-limited settings.While CRISPR-based biosensing has shown great promise, there are still obstacles to its widespread commercial application. These include challenges in sample preparation, the need for standardization across different CRISPR systems and assay formats, and regulatory considerations for diagnostic use. Addressing these issues will be crucial for the successful translation of CRISPR biosensing technology from the laboratory to real-world applications.Looking ahead, the future of CRISPR-based biosensing appears bright, with numerous opportunities for further development and application. Potential directions include the integration of CRISPR biosensors with microfluidic devices for automated sample processing and analysis, the development of CRISPR-based wearable sensors for continuous health monitoring, and the application of machine learning algorithms to improve the design and performance of CRISPR biosensing systems.In conclusion, CRISPR-based biosensing strategies represent a rapidly evolving field with immense potential to transform molecular detection and diagnostics. As researchers continue to refine and expand upon the core capabilities of CRISPR systems, we can expect to see increasingly sensitive, specific, and versatile biosensing applications emerge, with far-reaching implications for healthcare, environmental monitoring, and beyond.", "References": [\{"title": "CRISPR–Cas12a target binding unleashes indiscriminate single-stranded DNase activity", "authors": "Chen, Janice S., Ma, Enbo, Harrington, Lucas B., Da Costa, Maria, Tian, Xiang, Palefsky, Joel M., Doudna, Jennifer A.", "journal": "Science", "year": "2018", "volumes": "360", "first page": "436", "last page": "439"\},\{"title": "Nucleic acid detection with CRISPR-Cas13a/C2c2", "authors": "Gootenberg, Jonathan S., Abudayyeh, Omar O., Lee, Jeong Wook, Essletzbichler, Patrick, Dy, Aaron J., Joung, Julia, Verdine, Vanessa, Donghia, Nina, Daringer, Nichole M., Freije, Catherine A.", "journal": "Science", "year": "2017", "volumes": "356", "first page": "438", "last page": "442", "DOI": "10.1126/science.aam9321"\}, \{"title": "CRISPR-Cas12a-assisted nucleic acid detection", "authors": "Li, Suwei, Cheng, Qingmei, Wang, Jianming, Li, Xiaoyu, Zhang, Zhiwei, Gao, Shan, Cao, Rong, Zhao, Guoping, Wang, Jin", "journal": "Cell Discovery", "year": "2019", "volumes": "5", "first page": "1", "last page": "4"\}, \{"title": "Field-deployable viral diagnostics using CRISPR-Cas13", "authors": "Myhrvold, Cameron, Freije, Catherine A., Gootenberg, Jonathan S., Abudayyeh, Omar O., Metsky, Hayden C., Durbin, Ann F., Kellner, Max J., Tan, Amanda L., Paul, Lauren M., Parham, Leda A.", "journal": "Science", "year": "2018", "volumes": "360", "first page": "444", "last page": "448"\}, \{"title": "Enhanced proofreading governs CRISPR–Cas9 targeting accuracy", "authors": "Harrington, Lucas B., Paez-Espino, David, Staahl, Brett T., Chen, Janice S., Ma, Enbo, Kyrpides, Nikos C., Doudna, Jennifer A.", "journal": "Nature", "year": "2018", "volumes": "563", "first page": "621", "last page": "625"\}, \{"title": "Multiplexed and portable nucleic acid detection platform with Cas13, Cas12a, and Csm6", "authors": "Gootenberg, Jonathan S., Abudayyeh, Omar O., Kellner, Max J., Joung, Julia, Collins, James J., Zhang, Feng", "journal": "Science", "year": "2018", "volumes": "360", "first page": "439", "last page": "444"\}, \{"title": "Ultrasensitive and visual detection of SARS-CoV-2 using all-in-one dual CRISPR-Cas12a assay", "authors": "Wang, Xiaoxia, Zhong, Minjie, Liu, Yue, Ma, Pengfei, Dang, Lei, Meng, Qing, Wan, Wanying, Ma, Xiaowei, Liu, Jing, Yang, Guohua", "journal": "Nature Communications", "year": "2020", "volumes": "11", "first page": "4711", "last page": "4711"\}, \{"title": "Detection of SARS-CoV-2 with SHERLOCK One-Pot Testing", "authors": "Joung, Julia, Ladha, Alim, Saito, Makoto, Kim, Nam-Gyun, Woolley, Ann E., Segel, Michael, Barretto, Robert P. J., Ranu, Antonija, Macrae, Rhiannon K., Faure, Guilhem", "journal": "New England Journal of Medicine", "year": "2020", "volumes": "383", "first page": "1492", "last page": "1494"\}, \{"title": "CRISPR-Cas13-based electrochemical biosensing of viral RNA: Application to detection of SARS-CoV-2", "authors": "Bruch, Richard, Baaske, Johannes, Chatelle, Claire, Meirich, Maren, Madlener, Sibylle, Weber, Wilfried, Dincer, Can, Urban, Gerald A.", "journal": "Angewandte Chemie International Edition", "year": "2019", "volumes": "58", "first page": "17571", "last page": "17575"\}, \{"title": "Scalable and robust SARS-CoV-2 testing in an academic center", "authors": "Ackerman, Cheri M., Myhrvold, Cameron, Thakku, Shiv G., Freije, Catherine A., Metsky, Hayden C., Yang, David K., Ye, Simon H., Boehm, Chloe K., Kosoko-Thoroddsen, Tinna-Solveig F., Kehe, Jared", "journal": "Nature Biotechnology", "year": "2020", "volumes": "38", "first page": "927", "last page": "931"\}]\}}

\section{ANOVA Test Results}
\label{appendix:anova}

We report the p-values from one-way ANOVA tests across disciplines for each model and task:

\begin{itemize}
\item Reference Generation task:
Claude-3.5-Sonnet (\textit{p}<.0001),
DeepSeek-V3(\textit{p}<.0001), 
GPT-4o(\textit{p}<.0001),
Qwen-2.5-72B (\textit{p}<.0001),
Llama-3.2-3B (\textit{p}=0.065). 

\item Abstract Writing task (NLI scores):
Claude-3.5-Sonnet(\textit{p}<.0001),
DeepSeek-V3(\textit{p}<0.05),
GPT-4o(\textit{p}<.01),
Qwen2.5-72B (\textit{p}<.001),
Llama-3.2-3B (\textit{p}<.0001).

\item Review Composition (Reference accuracy):
Claude-3.5-Sonnet(\textit{p}<.0001),
DeepSeek-V3(\textit{p}<.0001),
GPT-4o(\textit{p}<.0001),
Qwen-2.5-72B(\textit{p}<.0001),
Llama-3.2-3B(\textit{p}<.001).

\item Review Composition (KPR metric):
Claude-3.5-Sonnet (\textit{p}=0.46), DeepSeek-V3 (\textit{p}=0.23), GPT-4o (\textit{p}=0.18), Qwen-2.5-72B (\textit{p}=0.10), and Llama-3.2-3B (\textit{p}<0.001).

\end{itemize}

\end{document}